# Quantitative methods for Phylogenetic Inference in Historical Linguistics: An experimental case study of South Central Dravidian

Taraka Rama, Sudheer Kolachina, Lakshmi Bai B.

**ABSTRACT**

In this paper we examine the usefulness of two classes of algorithms *Distance Methods, Discrete Character Methods* (Felsenstein and Felsenstein 2003) widely used in genetics, for predicting the family relationships among a set of related languages and therefore, diachronic language change. Applying these algorithms to the data on the numbers of shared cognates-with-change and changed as well as unchanged cognates for a group of six languages belonging to a Dravidian language sub-family given in Krishnamurti et al. (1983), we observed that the resultant phylogenetic trees are largely in agreement with the linguistic family tree constructed using the comparative method of reconstruction with only a few minor differences. Furthermore, we studied these minor differences and found that they were cases of genuine ambiguity even for a well-trained historical linguist. We evaluated the trees obtained through our experiments using a well-defined criterion and report the results here. We finally conclude that quantitative methods like the ones we examined are quite useful in predicting family relationships among languages. In addition, we conclude that a modest degree of confidence attached to the intuition that there could indeed exist a parallelism between the processes of linguistic and genetic change is not totally misplaced.

## 1 Introduction

Ever since the beginning of evolutionary thought, intuitions have been galore about the relevance of the process of evolution to Language change. In the field of linguistic theory itself, the idea of "a common origin" had existed long before Darwin's observation of 'curious parallels' between the processes of biological and linguistic evolution. The birth of comparative philology as a methodology is often attributed to that now very well-known observation by Sir William Jones that there existed numerous similarities between far-removed languages such as Sanskrit, Greek, Celtic, Gothic and Latin which was impossible unless they had 'sprung from some common source, which perhaps no longer exists'. This observation also marked the birth of the Indo-European language family hypothesis. Though Jones may not have been the first to suggest a link between Sanskrit and some of the European languages, it was only after his famous remarks that explanations for the enormous synchronic diversity of language started assuming a historical character. Up until that point in linguistic theory, explanations for the similarity and therefore, the relationship between different languages had been purely taxonomic and essentially ahistorical. See Atkinson and Gray (2005) for a very interesting and comprehensive comparative study of the historical development of linguistic and biological theory.

In the history of its development, the field of linguistics in general has crossed paths with biology on more than one occasion. One of the most significant interactions between these two disciplines witnessed the emergence of the biological nativist school of thought in the last century. Language came to be seen as a biological system rather than being a cultural artefact alone. Subsequent to this development but perhaps not directly related to it, Language became the object of study of a specialized area that has been referred to as Natural Signal Processing in the field of computer science. Attempts to discover the models underlying these natural systems were made as part of what was known as model-based analyses. The work that followed during this period saw a drastic cross-pollination of ideas across different domains. Computational/Quantitative methods proven to be useful in the analysis of a particular natural signal were applied to other natural signals and the results were studied. In recent times, one such attempt at cross-pollination has been the application of quantitative methods developed in the fields of Bioinformatics and Computational biology to language data. The intuition is that these methods which infer genetic phylogeny (evolutionary tree) from surface gene sequence data can do so from language data too. This would amount to saying that the model underlying genetic change is similar to the one underlying language change. In this paper, we survey some of these quantitative methods and test their performance for inferring the language tree of the South-Central Dravidian (formerly) sub-family from surface data. The goal of our study is to validate the use of such methods in Historical linguistics via a comparative study of the trees inferred computationally from surface data and the standard linguistic tree constructed on the basis of various phonological and morphological isoglosses.

The outline of this paper is as follows. Section 2 gives the basics and background of various terms used in bioinformatics/computational biology for inferring phylogenetic trees and their relevance in historical linguistics. Section 3 describes the dataset used in our experiments. Sections 4 describes the distance methods and the results of our experiments. Section 5 describes the character based methods and the results. We discuss the trees resulting from our experiments and compare them against the linguistically constructed tree in section 6 which is followed by conclusion and future work.

## 2 Basics and Related Work

The first attempt to apply quantitative methods to the historical study of language was made by the linguist Swadesh (1952) when he developed glottochronology and lexicostatistics. However, divergence time estimates for languages made using the glottochronological method have come to be seen by linguists as unreliable for a host of reasons[1]. Although much hyped during the initial stages of its development, it was not long before historical linguists became disillusioned with this first attempt to employ quantitative methods in the study of language relationships. A widespread skepticism about the efficacy of quantitative models that predict linguistic relationships from lexical data has prevailed within the historical linguistic community ever since.

However, in recent years, a new set of quantitative techniques emerged in computational biology that

---

1   Lexicostatistical methods infer linguistic relationships based on the number of shared cognates among languages. The comparision is done using a basic meaning list which is supposed to be culture-free and universal and hence, resistant to borrowing and replacement. This list comprises concepts such as body parts, numerals, elements of nature, etc. The first step is to collect the commonly used words in each language for each of these universal concepts in this list. In the second step, within the set of words from all the languages corresponding to a concept, sets of possible cognates are identified. The cognacy judgements are made on the basis of systematic sound correspondences among the words of different languages. Known cases of borrowings are discarded from the list. In the third step, the distance between every pair of languages is calculated as the number of cognates shared by those two languages. By using techniques like UPGMA, a family tree for a set of languages can be constructed using all the pairwise distances. Glottochronology goes one step further and estimates the divergence times for each node in the family tree. It assumes that the rate of lexical replacement is constant for all languages at all times. This constant is called the glottochronological constant and its value was assumed to be fixed at 0.806 . Swadesh (1952) used the following formula to estimate the divergence times of the Amerindian language family

$$t = \frac{\log c}{2 \log r}$$

where $r$ is the glottochronological constant and  $c$  is the percentage of shared cognates.
The glottochronological method has been criticised for the following reasons. First, there is a loss of information when the character-state data is converted to percentage similarity scores. Second, the case of a language having multiple or no words is not handled. Third, the assumption of a universal rate constant is disputable as the rate of evolution across languages has been observed to be variable. Fourth, the UPGMA method based on the percentage of shared cognates can produce inaccurate branch lengths and thus, lead to erroneous divergence times. In addition, the glottochronological method does not address the phenomena of reticulate evolution and parallel development at all. For these reasons, historical linguists disapprove of glottochronology as a valid method for the diachronic study of language.

could infer genetic phylogeny from gene sequence data. Researchers soon realized that these methods could be applied to language data too. Languages like genetic taxa shared changed traits (changed cognates) and language data like genetic data could be represented as state sequences. All of this resulted in a renewed interest in the application of quantitative methods to language data. The availability of data sets for well-established language families like Indo-European (Dyen et al. 1992) has spurred a number of researchers to apply these methods to these data sets and validate the resultant phylogenetic trees against the well-established linguistic facts and to test competing hypotheses. These methods are of two types: character based and distance based. We give an overview of the basic terminology in the following section.

## 2.1 Terminology

### 2.1.1 Characters

Language evolution can be seen as a change in some of its features. A *character* encodes the similarity between languages based on the values of these features and defines an equivalence relation on the set of languages $L$. Defining a character formally (Ringe et al. 2002)

A character is a function $c : L \rightarrow Z$ where $L$ is the set of languages and $Z$ is the set of integers.

A character can take different values across a set of languages indicating that these languages have different "states" with respect to that character. Two languages would have the same state with respect to a particular character if they have the same value for that character. The actual values of these characters are not important (Ringe et al. 2002).

Characters can either be lexical, phonological or morphological features. A **lexical** character corresponds to a meaning slot. For a given meaning, lexical items from different languages fall into different cognate classes (based on the cognacy judgments) which are then represented by different states for that lexical character. Two languages would have the same state if they have lexical items which are cognates. Figure 1 shows an example of how lexical characters are represented as states. The superscript shows the state exhibited by each language for a particular meaning slot.

| English | here$^1$ | sea$^5$ | water$^9$ | when$^{12}$ |
| --- | --- | --- | --- | --- |
| German | hier$^1$ | See$^5$, Meer$^6$ | Wasser$^9$ | wann$^{12}$ |
| French | ici$^2$ | mer$^6$ | eau$^{10}$ | quand$^{12}$ |
| Italian | qui$^2$, qua$^2$ | mare$^6$ | acqua$^{10}$ | quando$^{12}$ |
| Modern Greek | edo$^3$ | thalassa$^7$ | nero$^{11}$ | pote$^{12}$ |
| Hittite | ka$^4$ | aruna-$^8$ | watar$^9$ | kuwapi$^{12}$ |

FIGURE 1: An excerpt from the Dyen's Comparative Indo-European database

**Morphological** characters are normally inflectional markers and like lexical items, are coded based on cognation. **Phonological** characters are used to represent the presence or absence of a particular sound change(or a series of sound changes) in a set of languages.

### 2.1.2 Homoplasy and Perfect Phylogenies

Two languages can share the same state not only due to shared evolution but also due to phenomena of **backmutation** and **parallel development**. These phenomena are jointly referred to as **homoplasy**. For a particular character, if an already observed state reappears in the tree, then the phenomenon is called backmutation. Two languages may evolve independently in a similar fashion. In that case the two languages exhibit the same state despite evolving independently. This phenomenon is known as parallel development. Much of the related work in this area was based on the assumption of a homoplasy-free evolution (Ringe et al. 2002, Nakhleh et al. 2005b,a). When a character evolves without homoplasy down the tree then it is said to be *compatible* for that tree and the tree is said to be a **perfect phylogeny**. Hence, everytime a character's state changes in the tree, all the subtrees rooted at that point share the same state. Another source of ambiguity in the states of a character can be due to borrowing and is normally avoided by discarding all known cases of borrowings.

**2.2 Related Work**

The fashion in which characters evolve down the tree is described by a model of evolution. This specification or non-specification of models of evolution broadly divides the phylogenetic inference methods into two categories. For example, methods such as *Maximum Parsimony*, *Maximum Compatibility* and Distance methods such as *Neighbour Joining* and *UPGMA* do not require an explicit model of evolution. But other statistical methods like *Maximum Likehood* and *Bayesian Inference* are parametric methods which assume a model of evolution. The parameters of the model are tree topology, branch length and the rates of variation across characters. An interesting debate is going on in the scientific community regarding the appropriateness of the assumption of a model of evolution for linguistic data (Evans et al. 2004).

Gray and Jordan (2000) were among the first to apply the Maximum Parsimony method to Austronesian language data. They applied the technique to 5,185 lexical items from 77 Austronesian languages and were able to get a single most parsimonious tree. The maximum parsimony method returns that tree on which the minimum number of character state changes have taken place. There are different types of parsimonies such as Wagner, Camin-Sokal, etc and each of them makes different assumptions about the character state changes. These parsimonies are discussed in detail in section 5.

Of particular interest is the work of (Atkinson et al. 2005, Atkinson and Gray 2006) in which they applied Bayesian inference techniques (Huelsenbeck et al. 2001) to the Indo-European database. They used a matrix of binary values to represent the states of the languages for the lexical characters. Although their tree was identical to the tree established by the historical linguists using the comparative method[2], the dating based on *penalised likelihood* supported the famous Anatolian origin hypothesis as opposed to the Krugan hypothesis, dating the Indo-European family as being around 8000 years old. Their model assumes that the cognate sets evolve independently. They use a gamma distribution to model the variation across the cognate sets and try to find a sample of trees which best fits their data. Unlike the other non-parametric methods mentioned above, this method can handle polymorphism. By representing cognate information in terms of binary matrices (Figure 2), unlike glottochronology, there is no loss of information in this model. They also tested their model in scenarios where the cognacy judgements were not completely accurate and where the model misspecification could cause a bias in the estimate using an additional data set of ancient languages prepared by Ringe et al. (2002). They further tested their model on a synthetic data set which allowed for borrowing to occur between different lineages. The model was tested against two kinds of borrowing viz- borrowing between any two lineages and borrowing between lineages which are located locally. The dating in all the above cases was largely consistent with the dating they had obtained with the Dyen's dataset and this, they claim, demonstrates the robustness of the model.

Ryder (2006) in his work used syntactic features as characters and applied the above methods for inferring the phylogenetic tree for the Indo-European language family. He also used the same techniques on various language family data sets for grouping related languages into their respective language families. The syntactic features were obtained from the WALS database (Bakker 2004). The assumption underlying the use of syntactic features is that the rate at which syntactic features can be replaced is much less than that of

---

2   The position of Albanian is not resolved lexical features/items.

| Meaning | here | | | | sea | | | | water | | | when |
|---|---|---|---|---|---|---|---|---|---|---|---|---|
| Cognate set | 1 | 2 | 3 | 4 | 5 | 6 | 7 | 8 | 9 | 10 | 11 | 12 |
| English | 1 | 0 | 0 | 0 | 1 | 0 | 0 | 0 | 1 | 0 | 0 | 1 |
| German | 1 | 0 | 0 | 0 | 1 | 1 | 0 | 0 | 1 | 0 | 0 | 1 |
| French | 0 | 1 | 0 | 0 | 0 | 1 | 0 | 0 | 0 | 1 | 0 | 1 |
| Italian | 0 | 1 | 0 | 0 | 0 | 1 | 0 | 0 | 0 | 1 | 0 | 1 |
| Greek | 0 | 0 | 1 | 0 | 0 | 0 | 1 | 0 | 0 | 0 | 1 | 1 |
| Hittite | 0 | 0 | 0 | 1 | 0 | 0 | 0 | 1 | 1 | 0 | 0 | 1 |

FIGURE 2: An example of the binary matrix used by Atkinson and Gray (2006).

Ringe et al. (2002) proposed a computational technique called **Maximum Compatibility** for constructing phylogenetic trees. This technique seeks to find the tree on which the highest number of characters are compatible. Their model assumes that the lexical data is free of **back mutation and parallel development**. The method was applied to a data set of 24 ancient and modern Indo-European languages. They use morphological, lexical and phonological characters to infer the phylogeny of these languages. Nakhleh et al. (2005a) proposed an extension to this method known as **Perfect Phylogenetic Networks** which models homoplasy and borrowing explicitly. For a comparison of the various phylogenetic methods on the ancient Indo-European data, refer (Nakhleh et al. 2005b). They observed that almost all the methods, except UPGMA, produced trees which although partly similar, were strikingly different from one another in several ways. It must be noted that these scholars did not seek answers to much-disputed questions in the literature on the Indo-European language family tree, such as the status of Albanian in their afore-mentioned quantitative analyses. In each of the attempts discussed till now, the main thrust has been to demonstrate that language phylogeny as inferred using these quantitative methods was in almost perfect agreement with the traditional comparative method-based family tree thus demonstrating the utility of quantitative methods for the study of language change.

An earlier attempt to apply quantitative methods to the Dravidian language family data was made by Andronov (1964) using glottochronology. His dating of the Dravidian language family divergences was criticised for the largely faulty data used by him as it made the dating unreliable and untenable (Krishnamurti 2003). In other related work on the application of quantitative methods to the Dravidian language family, Krishnamurti et al. (1983) used unchanged cognates as a criterion for the subgrouping of South-Central Dravidian languages. Krishnamurti (1978) prepared a list of 63 cognates in all the six languages which he determined would be sufficient for inferring the language tree of the family within the framework of lexical diffusion. They examined a total of 945 rooted binary trees[3] and scored the trees based on the number of changes for all the 63 entries. The tree with the least score was the considered the best. Another related attempt was made by D'Andrade (1978) applying a technique called U-statistic hierarchical clustering to the same data set. In both these attempts, the trees obtained using the quantitative method/criterion were compared against the standard Dravidian language family tree constructed by Prof. Krishnamurti (2003) using the classical comparative method. We reproduce this widely-accepted standard tree here in Fig 3.

---

[3] $(2n-3)/2^{n-2}(n-2)!$

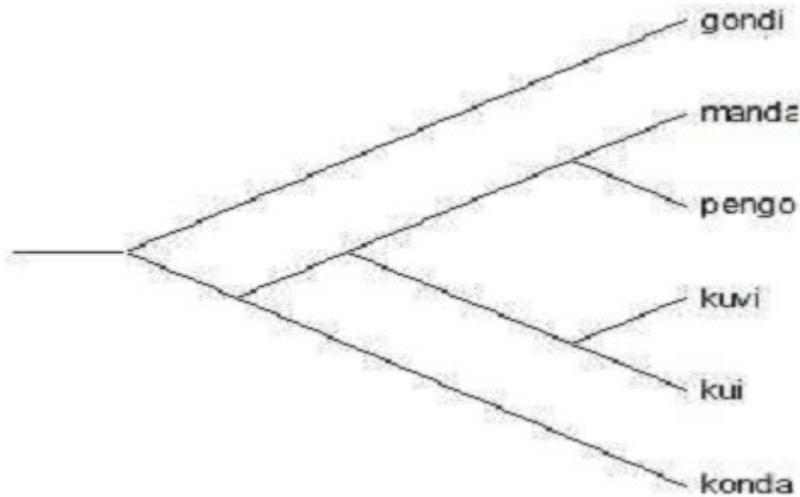

FIGURE 3: Tree constructed using Comparative Method

In our work, we deliberately avoided the use of lexical characters as they are noise-prone although much of the recent work in this area uses lexical characters for phylogenetic inference. Since the goal of our study was to evaluate the usefulness/performance of two classes of computational methods for historical linguistic tasks, we had to ensure that the data set used in the experiments was highly reliable for the evaluation to be accurate. This is why we chose to work with the same set of languages that Krishnamurti et al. (1983) studied. We converted the list of 63 changed/unchanged cognates given in that paper into phonological character data by treating each cognate as a character. By comparing the trees obtained by applying the character-based methods to this phonological character data with the standard tree in Fig 3, we verify the usefulness of these methods for phylogenetic inference.

**3 Dataset**

We used two different sets of data in our experiments. The data sets contain data from six South-Central (Now referred to as South Dravidian II in the recent literature. Refer to Krishnamurti (2003).) group of Dravidian Languages - viz. Gondi, Koṇḍa, Kui, Kuvi, Pengo, Manḍa. The first data set is the number of pair-wise shared cognates-with-change for all language pairs shown below in matrix form. The distance-based methods take this matrix as their input (Taken from Krishnamurti et al. (1983)).

|       | Gondi | Koṇḍa | Kui | Kuvi | Pengo |
|-------|-------|-------|-----|------|-------|
| Koṇḍa | 16    |       |     |      |       |
| Kui   | 18    | 18    |     |      |       |
| Kuvi  | 22    | 20    | 88  |      |       |
| Pengo | 11    | 19    | 48  | 49   |       |
| Manḍa | 10    | 9     | 40  | 42   | 57    |

TABLE 1: Matrix of shared cognates-with-change

The second data set was taken from Krishnamurti 1983 who provided a list of 63 cognate items which were qualified for study within the lexical diffusion framework. For each of these items, we represented languages with unchanged cognate as having state '0' and those with changed cognates as having state '1'. Thus we were able to represent this changed/unchanged cognate data as phonological characters with binary states.

**4 Distance Methods**

Distance-based methods take as input a matrix containing all the pair-wise distances for a given set of taxa (in this case, a set of languages) and output the phylogenetic tree which explains the data. The assumption of a lexical clock may or may not be required depending upon the method. In our study we examined two such methods which are very popular in evolutionary biology and have also been widely applied to historical linguistic data.

**UPGMA (Unweighted Pair Group Method with Arithmetic Mean)**
The lexicostatistical experiment for IE languages by Dyen et al. (1992) used this method for the construction of the family tree. This method can be described by the following psuedo-code:

1. Find the two closest languages (L1, L2) based on percentage of shared cognates.

2. Make L1,L2 siblings.

3. Remove one of them, say L1 from the set.

4. Recursively construct the tree on the remaining languages.

5. Make L1 the sibling of L2 in the final tree.

UPGMA assumes a uniform rate of evolution throughout the tree i.e, the distance of the root node to the leaves is equal. This algorithm produces a rooted tree whose ancestor (root) is known.

**Neighbour Joining (NJ)**

Neighbour Joining is a type of agglomerative clustering developed by Saitou (1987). It is a greedy method like UPGMA but unlike UPGMA it does not assume a uniform lexical clock. In addition, this method produces an unrooted tree along with the branch lengths which needs to be rooted for inferring the ancestral states and the divergence times for the languages. The method starts out with a star-like topology and then tries to minimize the estimate of the total length of the tree by combining together the languages that provide the maximum reduction. It has been mathematically shown that this method is statistically consistent (In other words, if there exists a tree which fits the lexical data perfectly, this method retrieves that tree). There are other distance-based methods such as FITSCH and KITSCH (Felsenstein and Felsenstein 2003) which are relatives (generalised versions) of UPGMA and NJ but we do not discuss them in this paper. A general observation about the class of distance-based methods is that the Neighbour Joining method outperforms all other methods in this class.

**4.1 Results for distance methods**

The tree structure in figure 3 is the South-Central Dravidian (South Dravidian II) sub-family tree constructed by Krishnamurti based on various morphological and phonological isoglosses. The similarity matrix in table 1 is converted into a distance matrix using the formula $d = 1/s_{ij}, i \leq j$. Figures 4 and 5 show the trees obtained by applying UPGMA and NJ methods to this pair-wise distance data.

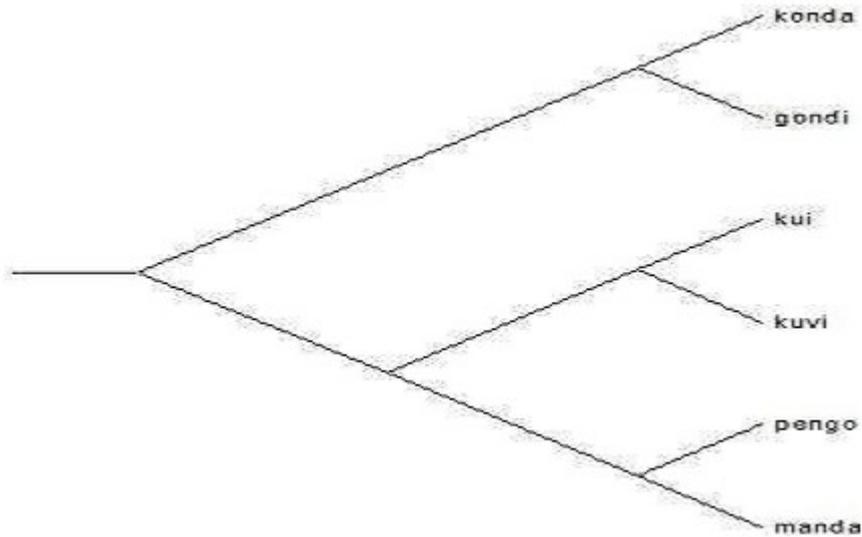

FIGURE 4: Phylogenetic tree using UPGMA

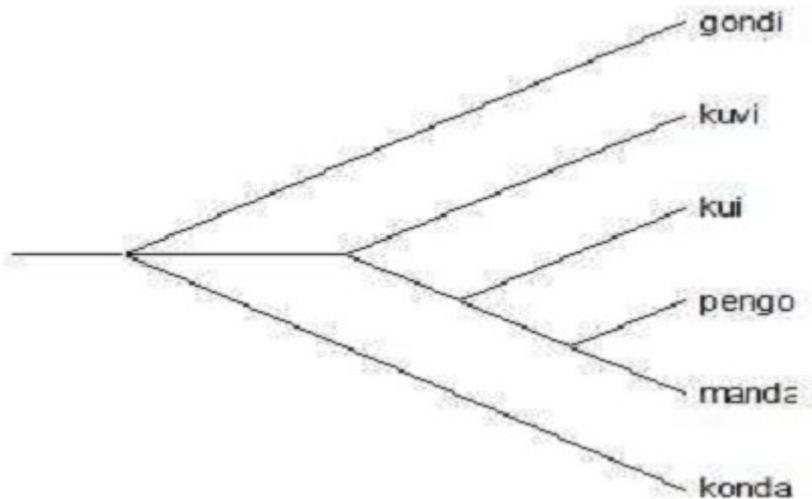

FIGURE 5: Phylogenetic tree using Neighbour Joining

## 5 Character Methods

**Maximum Parsimony**

With the exception of Bayesian analysis, Parsimonous methods have been claimed to be the most efficient for inferring the tree that is closest to the traditional standard tree (Ringe et al. 2006). There are various types of parsimonies depending upon the number of states (binary or multi-state) and the kind of transitions between the states. In our study, we limit ourselves to three kinds of parsimonies- Camin-Sokal, Wagner and Dollo parsimony. We reproduce the assumptions of each of these parsimonies as given in (Felsenstein and Felsenstein 2003) here.

Assumptions of Camin-Sokal's and Wagner's parsimony
1. Ancestral states are known (Camin-Sokal) or unknown (Wagner).

2. Different characters evolve independently.

3. Different lineages evolve independently.

4. Changes 0 → 1 are much more probable than changes 1 → 0 (Camin-Sokal) or equally probable (Wagner).

5. Both these kinds of changes are a priori improbable over the evolutionary time spans involved in the differentiation of the group in question.

6. Other kinds of evolutionary events such as retention of polymorphism are far less probable than 0 → 1 changes.

7. Rates of evolution in different lineages are sufficiently low so that two changes in a long segment of the tree are far less probable than one change in a short segment.

We also tested the effect of the hypothesis of the irreversibility of sound change by giving equal chance to change in both the directions. While Camin-Sokal parsimony corresponds to the case of sound change being irreversible, Wagner parsimony on the other hand, assigns equal probability to the occurrence of a sound change in both the directions.

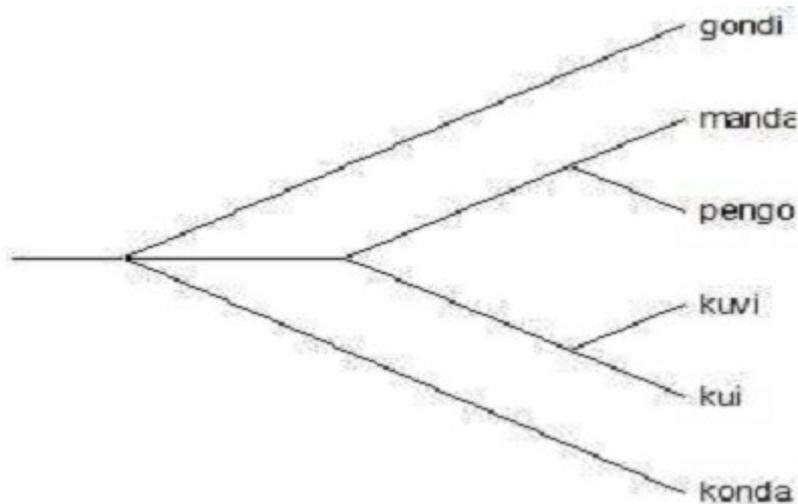

FIGURE 6: Phylogenetic tree using Wagner's parsimony

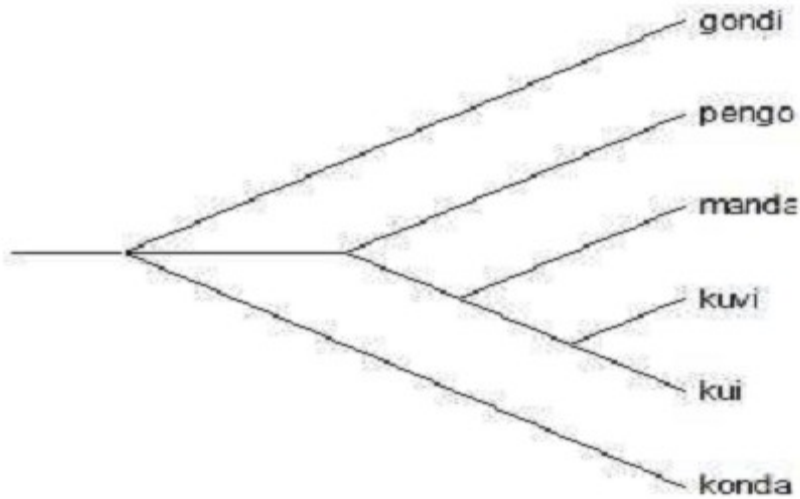

FIGURE 7: Phylogenetic tree using Wagner's parsimony

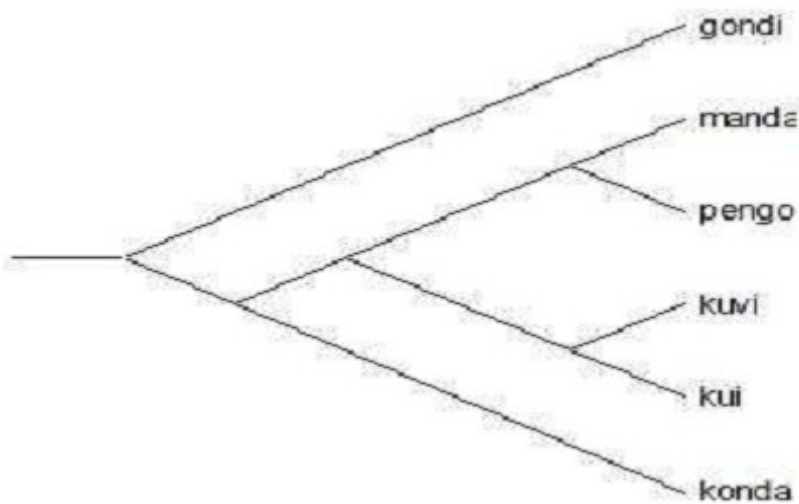

FIGURE 8: Phylogenetic tree using Camin-Sokal parsimony

Assumptions of Dollo's Parsimony
1. We know which state is the ancestral one (state 0).

2. The characters evolve independently.

3. Different lineages evolve independently.

4. The probability of a forward change ($0 \rightarrow 1$) is small over the evolutionary times involved.

5. The probability of a reversion ($1 \rightarrow 0$) is also small, but still far larger than the probability of a forward change, so that many reversions are easier to envisage than even one extra forward change.

6. Retention of polymorphism for both states (0 and 1) is highly improbable.

7. The lengths of the segments of the true tree are not so unequal that two changes in a long segment are as probable as one in a short segment.

Dollo's parsimony is based on Dollo's law which states that traits can evolve only once. In this context, the evidence of cognates which represent the process of diffusion of sound change still in process, can be treated as trait. This is equivalent to stating that sound change is homoplasy free. In other words, Sound change has diffused across the languages at a common stage in their evolution rather than occur at a later stage when the languages have diverged. This kind of parsimony also allows for determining the root of the tree.

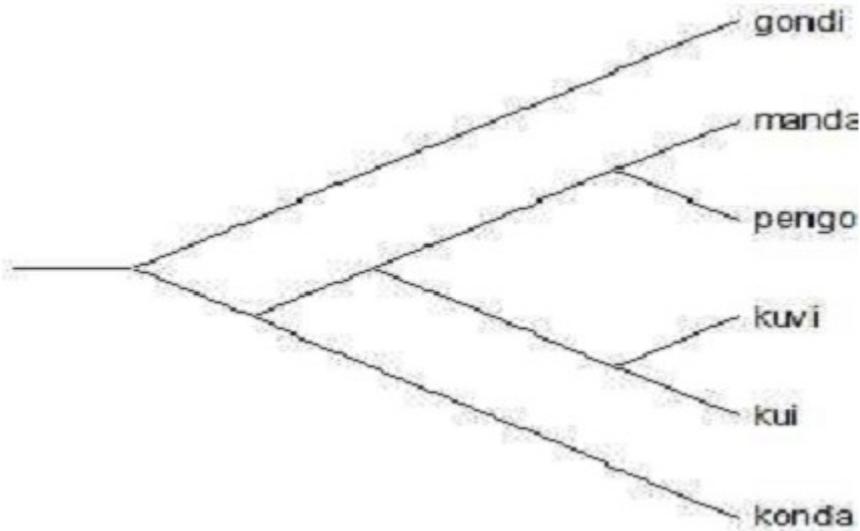

FIGURE 9: Phylogenetic tree using Dollo's parsimony

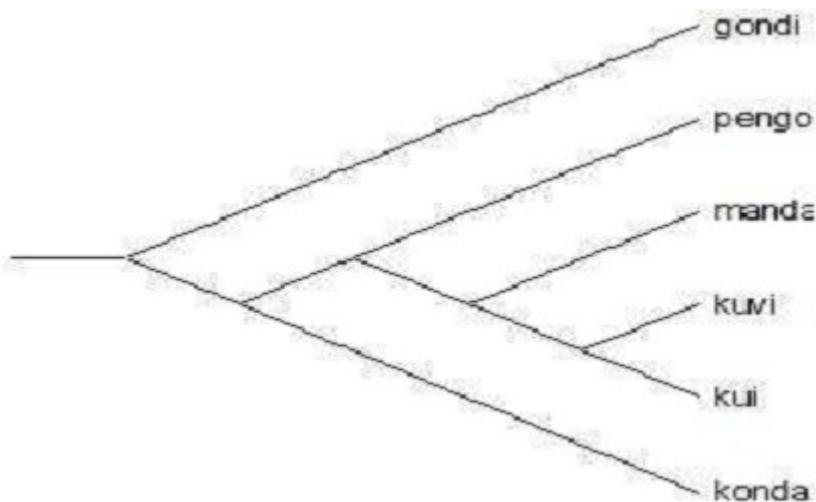

FIGURE 10: Phylogenetic tree using Dollo's parsimony

**Bayesian Inference of Phylogenies**
This is a different class of character-based methods which is an extension of the maximum likelihood method. We used Metropolis-coupled Markov Chain Monte Carlo (MCMC) for sampling the posterior probabilities of the trees. This method was employed by Atkinson et al. (2005) and has been discussed in some detail in the related Work section. We experimented with different parameter settings and observed their effect on the inferred trees. We tried using two priors, one a fixed shape parameter (α) and the other a uniform distribution. The results did not vary much when we changed the priors. MCMC runs *n* chains out of which *n*-1 chains are heated. A heated chain has the steady-state distribution $\pi_i(X) = \pi(X)^{\beta_i}$ with $\beta_i = \frac{1}{1+T(i-1)}$ where *T* is the temperature, *i* is the number of the chains, π is the posterior distribution and *β* is the power to which the posterior probability of each heated chain is raised. The chains are heated in an incremental fashion and after each iteration, the states of two randomly picked chains *i* and *j* are swapped with the following probability

$$\pi_i(X_i) = min\left(1, \frac{\pi_i(X_t^{(j)})\pi_j(X_t^{(i)})}{\pi_i(X_t^{(i)})\pi_j(X_t^{(j)})}\right)$$

Inferences or sampling is usually done on the cold chain with *β* = 1 and *T* = 0.20 and the number of chains *n* = 4 . We ran two independent analyses and the chains were kept running until the average deviation of the split frequencies between the two analyses was less than 0.01. The first 25% of the analyses were thrown out as part of the burn-in.

**6 Discussion**

We compared the trees resulting from our experiments with the traditional tree topology given by Krishnamurti. First we discuss the trees obtained using distance-based methods which take the number of shared cognates (table 1) as input. To our surprise, UPGMA which is less sophisticated outputted the tree that best matches the data in table 1. In his 1983 paper, Krishnamurti briefly cites the issues that were pointed out to him in the Fig 3. The tree makes 40 predictions out of which 37 are correct and 3 are wrong. The wrong predictions are 1) Kuvi should be closer to Koṇḍa than it is to Gondi. This prediction is wrong as Kuvi shares 20 innovative items with Koṇḍa but 22 with Gondi 2) Koṇḍa should be closer to Manḍa than it is to Gondi. However, Koṇḍa shares only 9 items with Manḍa but as many as 16 items with Gondi which makes this prediction wrong. 3) Manḍa should be closer to Koṇḍa than it is to Gondi. This last prediction also turns out to be wrong since Manḍa shares 10 items with Gondi but only 9 items with Gondi. All these wrong predictions are absent in the tree given by UPGMA as Gondi and Koṇḍa are placed under the same subtree. The other 37 correct predictions were not listed by Krishnamurti and hence could not verified in the UPGMA tree. However, as the UPGMA tree excepting the placement of Gondi and Koṇḍa is identical to the tree given by Krishnamurti, there is good reason to believe that the UPGMA tree is also able to succesfully make these correct predictions. Interestingly, the other distance-based method which is the neighbour joining method infers a tree which is identical to the one obtained by Krishnamurti using two sound changes[4]. Neighbour joining method returned an unrooted tree which we rooted treating Gondi as the outgroup.

The results obtained in the next set of experiments using character-based methods on phonological character representation of unchanged cognates yielded trees which are largely in conformity with the standard tree in Fig 3. (Compare Figs 6, 7, 8, 9 with Fig 3) We employed three different kinds of parsimony and each of them outputted similar trees. Wagner's and Dollo's parsimonies returned the two most parsimonious trees whereas Camin-Sokal's parsimony returned only one tree. The trees returned by Wagner's and Dollo's parsimonies are alike. Each of the parsimonious methods returned a tree which is identical to the standard tree constructed using the comparative method. The tree selected using Krishnamurti's scoring criterion and the one returned by the Camin-Sokal parsimony are identical. The lowest scoring tree selected using Krishnamurti's scoring criterion has a score of 71 and so do all the trees returned by these various parsimonies. Both Wagner's and Dollo's parsimonies returned an extra tree which had a score of 71. The extra tree returned by Wagner's and Dollo's parsimonies is actually ranked as the second best using Krishnamurti's scoring criterion. This is actually an important result as the relaxation of the irreversibility of sound change constraint gives two trees with the same score[5]. This indicates that as far

---
4   Please refer to the appendix of Krishnamurti's 1983 paper for the sound changes
5   Wagner's parsimony-Sound change is equiprobable in both directions as character-based quantitative methods are concerned, the irreversibility hypothesis can be safely dispensed with. In the case of Dollo's parsimony, the main assumption is that a new trait is very difficult to acquire but very easy to loose. In other words, it is easier for a language to revert to an older sound than to undergo a new sound change. This method also returned an extra tree which is identical to the one ranked second by Krishnamurti.

In fact, Krishnamurti's intuitive tree selection criterion developed in his 1983 work is a precursor to the various parsimony methods that have become so popular in different areas of research. Krishnamurti's assumptions are similar to the assumptions made by Camin-Sokal parsimony. We used Camin-Sokal parsimony to score the tree returned by UPGMA and obtained a score of 79. Examining the trees returned by the Bayesian analysis, we observed that it essentially returns a tree identical to the one returned by the neighbour joining method but with ternary branching- Gondi, Koṇḍa and the other languages as branches. There is a general ambiguity about the grouping of Manḍa and Pengo as well as that of Kui and Kuvi. All these issues surfaced in Krishnamurti et al (1983) too where the explanation put forth in section 4.3 was that this discrepancy resulted from the very nature of the sound change C2[6]. Thus, we observed that cases where the trees inferred using the quantitative methods differed from the standard tree constructed using the comaparative method, were in fact, cases of genuine ambiguity even to the historical linguist. The branch lengths returned by all the methods agree upon the fact that Gondi diverged earlier than the other languages and is followed by Koṇḍa.

**7 Conclusion and Future Work**

To the best of our knowledge, this is the first time ever that methods from bioinformatics and computational biology have been applied to Dravidian language data for phylogenetic inference. On the basis of a comparison of the resulting trees with the standard tree and also the trees found in other earlier work, we seek to evaluate the performance of these methods for phylogenetic inference. The trees inferred using the different quantitative methods are largely in agreement with the linguistic facts. Character-based methods outperformed the distance-based methods going by Krishnamurti's scoring criterion. In fact, the performance of the distance-based methods is itself quite decent. The UPGMA tree rightly rules out the wrong predictions cited by Krishnamurti. Since we ensured that the data was noise-free by using well-known data sets carefully prepared by an expert, we can unambiguously claim that the good performance of these quantitative methods is a reflection of their usefulness for the historical linguistic task of phylogenetic inference. In addition to correctly inferring the structure of the family tree, these methods have an added advantage of being able to return the branch lengths in a tree. These branch lengths can be used to calibrate the divergence times of the tree and can throw light upon the antiquity of the Dravidian language family although such dating should be done with sufficient attention to detail and more importantly, under the supervision of a trained linguist to prevent spurious dates. We could not address the issue of dating as we lack the required expertise. We hope that dates estimated using qunatitative methods such as the ones discussed in our work become a kind of starting point for a detailed investigation into the antiquity of the Dravidian language family. The results of our experiments also validate the hypothesis that Language being a natural system, the processes underlying it are the same as those underlying any other natural system and hence, there could exist a parallelism between language change and genetic change.

We conclude with the remark that quantitative methods such as the ones we surveyed can reliably infer language phylogeny and certainly merit consideration in Historical linguistics. However, it must also be mentioned that the phylogenetic trees outputted by these methods must not be treated as the final word on Language phylogeny and need to be linguistically verified at every stage. In fact, we would recommend that these methods be tried out by researchers in historical linguistics on different data sets under the close

---

[6] C2: CVL > CLV > LV.

supervision of experts so that their veracity can be better ascertained.

## Colophon

We are much grateful to Prof. Bh. Krishnamurti for pointing out his paper to us without which we could not have had the right data to conduct our experiments. We thank the following for their proofreading and helpful comments: Prasanth Kolachina, Akshat Kumar, Sarath Chandra Addepalli and Karthik Gali.